\documentclass{article}

\usepackage[preprint, nonatbib]{neurips_2025}

\usepackage[utf8]{inputenc} % allow utf-8 input
\usepackage[T1]{fontenc}    % use 8-bit T1 fonts
\usepackage{url}            % simple URL typesetting
\usepackage{booktabs}       % professional-quality tables
\usepackage{amsfonts}       % blackboard math symbols
\usepackage{nicefrac}       % compact symbols for 1/2, etc.
\usepackage{microtype}      % microtypography
\usepackage{xcolor}         % colors
\usepackage[pdftex]{graphicx}
\usepackage{algorithm}
\usepackage{algpseudocode}
\usepackage{amsmath}
\usepackage{siunitx}
\usepackage{vcell}
\usepackage{multirow}
\usepackage{hyperref}       % hyperlinks
\usepackage{enumitem}

\title{An LLM-based Framework for Human-Swarm Teaming Cognition in Disaster Search and Rescue}

\author{%
  Kailun Ji$^{1}$\thanks{Equal contribution.} \quad
  Xiaoyu Hu$^{1}$\footnotemark[1] \quad
  Xinyu Zhang$^{1,2}$\thanks{Corresponding author: \texttt{Xinyu.Zhang@nwpu.edu.cn}} \quad
  Jun Chen$^{1,2}$ \\
  \\
  $^{1}$School of Electronics and Information, Northwestern Polytechnical University, Xi’an 710072, China\\
  $^{2}$Chongqing Institute for Brain and Intelligence, Guangyang Bay Laboratory, Chongqing 400064, China
}

\begin{document}

\maketitle

\begin{abstract}
Large-scale disaster Search And Rescue (SAR) operations are persistently challenged by complex terrain and disrupted communications. 
While Unmanned Aerial Vehicle (UAV) swarms offer a promising solution for tasks like wide-area search and supply delivery, yet their effective coordination places a significant cognitive burden on human operators. 
The core human-machine collaboration bottleneck lies in the ``intention-to-action gap'', which is an error-prone process of translating a high-level rescue objective into a low-level swarm command under high intensity and pressure.
To bridge this gap, this study proposes a novel LLM-CRF system that leverages Large Language Models (LLMs) to model and augment human-swarm teaming cognition. 
The proposed framework initially captures the operator's intention through natural and multi-modal interactions with the device via voice or graphical annotations. 
It then employs the LLM as a cognitive engine to perform intention comprehension, hierarchical task decomposition, and mission planning for the UAV swarm. 
This closed-loop framework enables the swarm to act as a proactive partner, providing active feedback in real-time while reducing the need for manual monitoring and control, which considerably advances the efficacy of the SAR task.
We evaluate the proposed framework in a simulated SAR scenario. 
Experimental results demonstrate that, compared to traditional order and command-based interfaces, the proposed LLM-driven approach reduced task completion time by approximately $64.2\%$ and improved task success rate by $7\%$. 
It also leads to a considerable reduction in subjective cognitive workload, with NASA-TLX scores dropping by $42.9\%$. 
This work establishes the potential of LLMs to create more intuitive and effective human-swarm collaborations in high-stakes scenarios.
\end{abstract}

\section{Introduction}
In large-scale disaster scenarios, such as earthquakes, floods, and fires, to name a few, securing the ``golden $72$-hour'' rescue window is paramount for saving lives and reducing losses \cite{boroujeni2024comprehensive, sadrabadi2025conceptual, xu2024emergency}. 
Under such a condition, Unmanned Aerial Vehicle (UAV) swarms have emerged as a critical asset in this race against time.
They are capable of rapid deployment to high-risk and inaccessible areas, and they can collaboratively perform essential tasks, including wide-area search \cite{chen2025fusion, zhang2022webuav}, target identification \cite{liu2024shooting}, building damage assessment \cite{zhong2022multi, vieira2023insplad}, and emergency medical supply delivery \cite{wandelt2023aerial}. 
The powerful capabilities of drone swarms, however, introduce a significant operational bottleneck in the form of an immense cognitive workload for human operators. 
% Faced with complex and dynamic conditions like aftershocks, severe weather, or heavy mist, on-site personnel must make rapid decisions within intense time pressure and heavy task loads. 
This difficulty is further compounded by the need to process and fuse multi-source and heterogeneous information streams, such as real-time drone video feedback, infrared thermal imaging, Geographic Information System (GIS) map data, and survivor reports, which is a demanding task that requires operators to sustain a high level of situational awareness throughout the entire operation \cite{deng2023vr}.

Traditional human-swarm interaction, which operates on a ``command-response'' paradigm, exacerbates the operator's cognitive workload \cite{peng2021review, bu2024advancement}. 
This approach generally requires the operator to manually decompose high-level Search And Rescue (SAR) intentions into a lengthy series of low-level machine instructions. 
For instance, to execute an order such as ``immediately send two drones to the collapsed red-roofed building in area B to check for life signals, and have another drone provide a high-altitude communication relay''.
An operator must manually: $1$) identify the building's precise coordinates, $2$) plan individual obstacle avoidance routes for the two drones, $3$) configure their sensor payloads (e.g., activating thermal imaging), $4$) set loitering waypoints and altitude for the relay drone, and $5$) continuously monitor and intervene the devices in real-time. 
This manual ``intention-to-command'' translation is highly inefficient and error-prone under high pressure and high intensity tasks, creating a significant gap between human decision-making and machine execution. 

\begin{figure}[t]
    \centering
    \includegraphics[width=\linewidth]{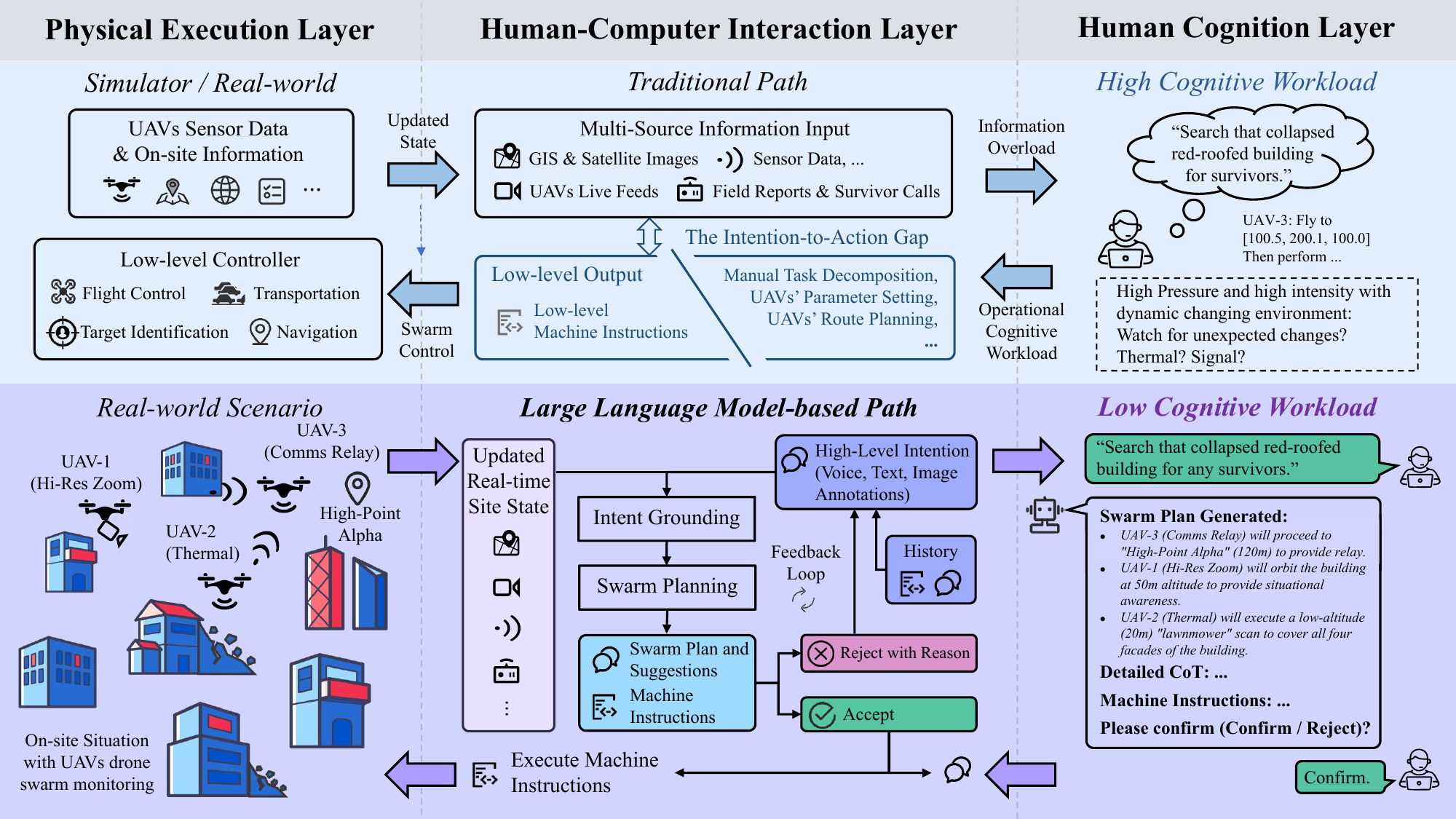}
    \caption{\textbf{The UAV Swarm Disaster SAR Workflow.} 
    The traditional approach (above) creates a significant ``intention-to-action gap'', imposing a heavy cognitive workload on human operators. 
    Our proposed framework (below) bridges this gap by leveraging an LLM-based core to intelligently decompose high-level multi-modal intention into an executable swarm plan.}
    \label{fig1}
\end{figure}

In addition, the process of translating human intention into commands is significantly influenced by both operator's preferences and mission-specific requirements, which directly shape the desired swarm behavior. 
For example, a wide-area search task prioritizes coverage efficiency, favoring a Z-shaped scanning pattern. 
Conversely, a building damage assessment demands high precision, for which a concentrated orbiting pattern is better suited. 
Traditional systems are inherently inflexible and cannot dynamically incorporate such implicit contextual knowledge or user-specific preferences, which limits their overall effectiveness and adaptability \cite{phung2021safety}.

Against this backdrop, Large Language Models (LLMs) demonstrate immense potential for addressing these bottlenecks \cite{brown2020language, chowdhery2023palm, chang2024survey}. 
The powerful capabilities of LLMs in natural language understanding, contextual reasoning \cite{kojima2022large, wei2022chain, feng2023towards}, multi-turn dialogue, and robust inference, have led to their effective applications in various complex human-machine collaboration scenarios, such as robotics (e.g., Google's SayCan)\cite{qiu2024dronegpt}, code generation (e.g., Github Copilot), and complex data analysis (e.g., OpenAI's Code Interpreter). 
The emergence of LLMs makes it possible for machines to comprehend high-level and ambiguous instructions. 
Therefore, this paper proposes a novel LLM-based cognitive reasoning framework (LLM-CRF) for human-swarm teaming. 
The proposed framework leverages the natural language understanding and reasoning capabilities of an LLM to model and augment the operator's cognitive processes. 
It captures high-level intention through multi-modal interactions (e.g., voice, gestures) \cite{li2021uav, sun2022human, ribeiro2021web} and employs the LLM as a central cognitive engine to autonomously perform intention comprehension, hierarchical task decomposition, and mission planning for the UAV swarm. 
By effectively closing the loop between human intention and swarm action, this approach transforms the UAV swarm from a passive tool into a proactive partner, thereby significantly advancing the efficacy of SAR operations.
The gap between the traditional and the proposed approaches is demonstrated in Figure \ref{fig1}.

\section{Related works}
Traditional research on UAV swarm coordination has predominantly addressed the algorithmic challenges of Multi-Agent Task Allocation (MATA) and Multi-Agent Path Planning (MAPP) \cite{peng2021review, bu2024advancement, ouyang2023formation}. 
This substantial body of research has yielded computationally efficient methods, including heuristic algorithms and market-based mechanisms to optimize swarm behavior for predefined objectives \cite{han2021modified, yan2024cooperative, skaltsis2023review, zhang2022dynamic}. 
However, these frameworks share a fundamental limitation in their underlying assumption, that a human operator can formally and precisely articulate mission goals, constraints, and cost functions in a structured, machine-readable format. 
This critical prerequisite, the ``intent-to-command'' translation, imposes a substantial cognitive burden on the operator \cite{deng2023vr, ribeiro2021web}. 
Consequently, the human capability is effectively reduced from a strategic commander to a low-level programmer, which is a role mismatch that proves particularly debilitating in the dynamic, high-stress environments characteristic of disaster response.

To mitigate such a burden, prior research has explored more intuitive interaction modalities, such as voice and gesture control \cite{divband2021designing}. 
These systems typically employ conventional Natural Language Processing (NLP) techniques, including semantic parsing and intention classification \cite{zhang2020rfhui}, to map a constrained vocabulary of predefined commands (e.g., ``take off'', ``scan area'') onto specific robotic functionalities. 
While representing a step forward, these approaches are inherently brittle and lack robustness, as they fail to comprehend the complex, contextual, and ambiguous instructions that typify real-world mission directives, such as ``check that collapsed red-roofed building for survivors''.
While the advent of LLMs \cite{brown2020language, chowdhery2023palm, vaswani2017attention} and Vision-Language Models (VLMs) \cite{liu2023visual, li2022blip, dai2023instructblip} has provided a transformative new path. 
Their powerful common-sense reasoning, in-context learning, and planning capabilities \cite{kojima2022large} have acted as the ``brain'' for embodied agents. 
Building on their success in robotics, LLMs have shown the capacity to ground high-level instructions into executable action sequences for manipulation and navigation \cite{vemprala2024chatgpt, qiu2024dronegpt}, and this paradigm is now being extended to UAVs. 
Recent research confirms that LLMs can effectively generate navigation waypoints, write flight control scripts, and perform high-level task planning for drone operations \cite{zhong2024safer, hu2022lora}.

Despite recent advances, current methods are not well-suited for disaster response, revealing several key shortcomings. 
A major limitation is that existing works overwhelmingly focus on single-agent control \cite{vemprala2024chatgpt, qiu2024dronegpt}, which does not address the one-to-many decomposition for swarm coordination.
This type of approach lacks the ability of complex resource allocation, role assignment, and spatio-temporal de-confliction. 
In addition, these systems are generally performed and evaluated in simple, structured, and simulated environments, which do not reflect the dynamic, unpredictable, and communication-constrained nature of a disaster site. 
Furthermore, the safety-critical context of SAR cannot tolerate the known risk of LLM hallucinations \cite{liu2024survey, favero2024multi}. 
Blindly executing a factually incorrect or made-up plan in a rescue mission is unacceptable, yet current frameworks lack the robust verification and human-in-the-loop feedback mechanisms required for such high-stakes operations. 
Hence, this work aims to tackle these concerns by creating a framework that can reliably and safely translate human commands into coordinated actions for drone swarms in real-world SAR missions.

\section{Method}
\begin{figure}[t]
    \centering
    \includegraphics[width=\linewidth]{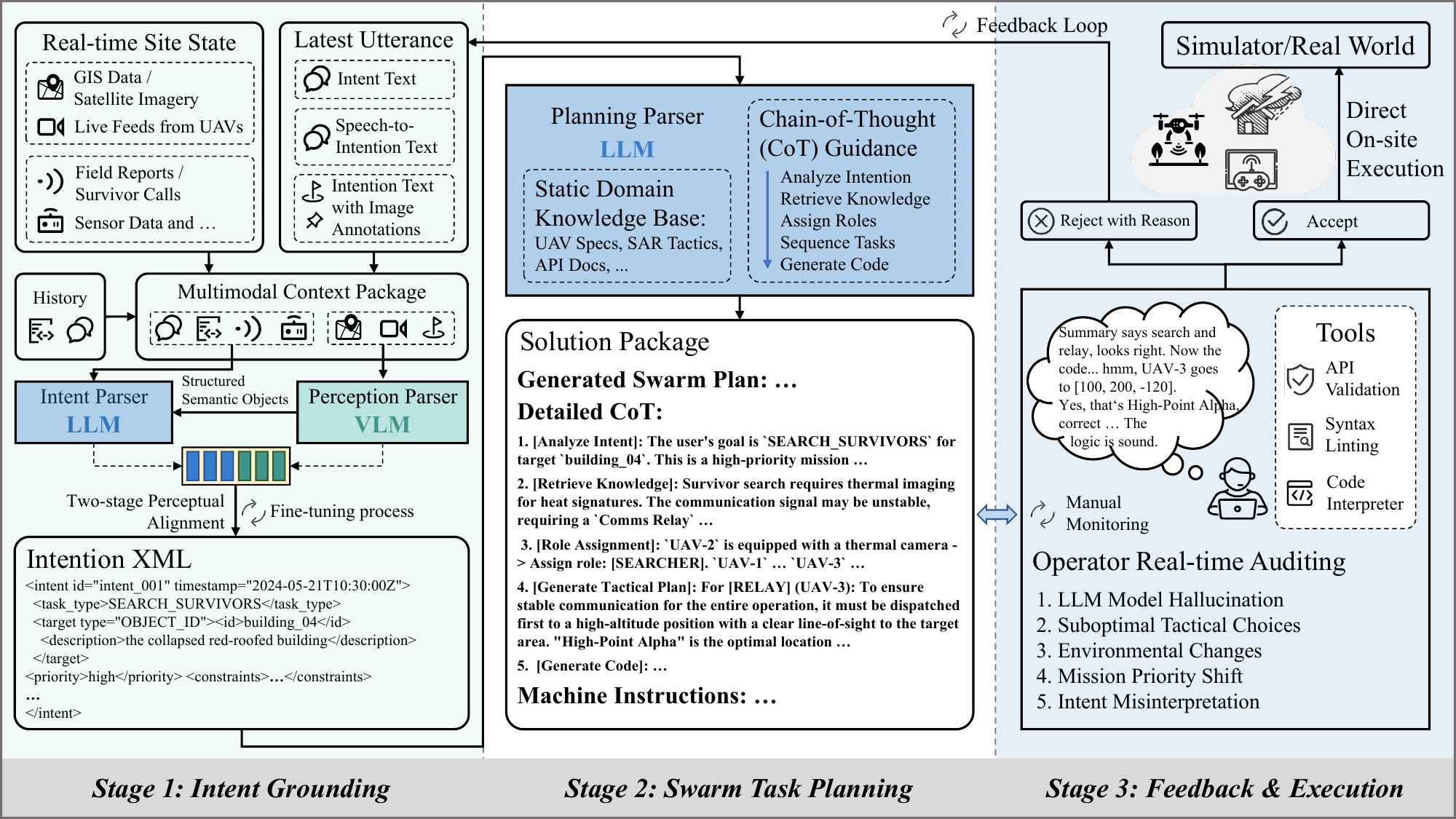}
    \caption{\textbf{The proposed LLM-based Cognitive Reasoning Framework (LLM-CRF).} 
    The system translates raw multi-modal inputs into executable actions through a three-stage process, including intent grounding, swarm task planning, and feedback and execution.}
    \label{fig2}
\end{figure}

This paper introduces a novel LLM-based Cognitive Reasoning Framework (LLM-CRF) to bridge the ``intention-to-action'' gap in UAV swarm control, translating high-level human intent into executable robotic actions. 
It functions as a cognitive engine between a front-end multi-modal interface and a back-end UAV action library, as illustrated in Figure \ref{fig2}. 

The LLM-CRF engine operates on a hierarchical, multi-model architecture. 
At its core, an LLM functions as the central reasoning agent, which is supported by a suite of specialized perception and transcription tools (i.e., Qwen-$14$B-Chat \cite{bai2023qwen}). 
Specifically, a Vision-Language Model (VLM) (i.e., LLaVA-$1.6$ \cite{liu2024llavanext}) is utilized as a dedicated visual perception module, while Whisper \cite{radford2023robust} serves as the speech-to-text module. 
Our methodology focuses on the deep, domain-specific adaptation of these models to form a cohesive expert system. 
The system initiates each decision cycle by constructing a multi-model context package. 
This package integrates all pertinent information for reasoning, including the operator's $latest\_utterance$, associated $image\_annotations$, the complete $dialogue\_history$, and the $world\_state$, describing the global site situation and UAV statuses. 
This consolidated package provides a unified input for all subsequent processing stages.

\subsection{Intent Grounding via Perceptual Alignment}
The initial stage of the LLM-CRF is Intent Grounding, which converts an operator's raw, multi-modal inputs into a structured and machine-executable representation. 
This requires the LLM to achieve a contextualized understanding of the disaster scene, semantically grounding the operator's linguistic commands within the UAVs' visual perceptions of the environment. 
To this end, we designed a two-stage Perceptual Alignment Fine-tuning process to address the domain shift problem when applying general-purpose VLMs to the specialized domain of UAV-based disaster response.
The fine-tuning is applied to achieve precise alignment between the VLM's representations and the LLM's semantic space, adapting it for the UAV-SAR domain through the following specialized stages:
\begin{itemize}[leftmargin=0.3cm]
\item Stage $1$ - Vision-language feature alignment pre-training:
This initial stage aims to establish a foundational mapping between visual features and general linguistic concepts. 
During this stage, we freeze the pre-trained vision encoder $f_{vision}(\cdot)$ and LLM $f_{llm}(\cdot)$, training only a lightweight adapter module $f_{adapter}(\cdot; \theta_{adapter})$. 
This adapter is trained on large-scale aerial image-text pairs ($I, T$) (e.g., from RS5M \cite{zhang2024rs5m}) to learn a projection that effectively maps visual features from the aerial domain into the LLM's embedding space. 
The optimization objective is to minimize the contrastive loss between the projected visual features and the text embeddings:
\begin{equation}
\label{eq:stage1}
\mathcal{L}_{stage1} = \mathcal{L}_{contrastive}(f_{adapter}(f_{vision}(I)), f_{llm}(T))
\end{equation}
This provides the model with a preliminary and generic understanding of visual semantics.

\item Stage $2$ - Domain-specific multi-modal instruction fine-tuning:
This stage elevates the model from a passive observer to an active perceptual agent within the disaster response context. 
Building upon the first stage, we continue to keep the vision encoder frozen while performing parameter-efficient fine-tuning via Low-Rank Adaptation (LoRA) \cite{hu2022lora}), on both the adapter and the LLM's parameters ($\theta_{adapter}$ and $\theta_{llm}^{LoRA}$).  
This stage is trained on a self-acquired multi-modal instruction dataset, which includes complex, scenario-specific tasks like Visual Question Answering (VQA). Given a visual input $I_d$ and a question $Q_d$ from the disaster domain, the model is trained to generate the correct answer $A_d$. The optimization objective is to minimize the standard language modeling (cross-entropy) loss:
\begin{equation}
\label{eq:stage2}
\mathcal{L}_{stage2} = -\sum_{i=1}^{|A_d|} \log P(A_{d,i} | I_d, Q_d, A_{d,<i}; \theta_{adapter}, \theta_{llm}^{LoRA})
\end{equation}
This training transforms the VLM from a passive image descriptor into an interactive perceptual module, capable of responding to and executing vision-grounded commands. 

\item Stage $3$ - Handling high resolution imagery:
To process detailed UAV imagery effectively, we employ a dynamic local feature perception strategy with a hybrid visual encoding.
It ensures the model to capture fine-grained local details while retaining crucial global contextual information during feature extraction.
\end{itemize}

During inference, this optimized VLM, denoted as $f_{vlm}(\cdot)$, functions as a dedicated perception module, processing real-time visual data $I_{stream}$ such as video streams into structured semantic objects $O_{sem}$ to update the $world\_state$. This can be formalized as:
\begin{equation}
\label{eq:inference_vlm}
O_{sem} = f_{vlm}(I_{stream}, \text{Detect all relevant object})
\end{equation} 

Subsequently, the core LLM, acting as the central reasoning agent, synthesizes this structured visual information with the broader context to form a coherent and actionable understanding of the operator's intention.
Guided by a structured prompt template, the LLM performs high-level semantic fusion, synthesizing all information to produce a standardized "Intention XML" representation. 
This XML schema is meticulously engineered to serve as a direct interface for the subsequent task planning stage. 
A typical output includes the following key fields: 
$<task\_type>$ defines the core action of the task, 
$<target>$ describes the specific object of the task (i.e., $OBJECT\_ID$, $COORDINATES$) and unique identifier, 
$<priority>$ is used for decision-making in case of resource conflicts, $<constraints>$ contains conditions or preferences that affect tactical choices (e.g., $use\_thermal\_imaging$), 
and $<spatial\_context>$ defines the geospatial scope of the task. 
This structured output provides a solid foundation for subsequent automated planning.

\subsection{Swarm Task Planning via In-Context Learning}
Upon accurately understanding the task objective through the intent grounding procedure, the this module is then responsible for decomposing the structured Intent XML into a multi-agent and parallelizable ``Solution Package''. 
A key challenge here is enabling a general-purpose LLM to execute the complex yet domain-specific reasoning this requires. 
Instead of the conventional path of fine-tuning on large-scale expert data, we employ a flexible In-Context Learning (ICL) strategy. 
The static knowledge base contains the necessary domain contextual and operational constraints, including UAV performance parameters, standard SAR tactics, and API specifications. 
A key advantage of this architecture is its flexibility, where the knowledge base can be modified and expanded without requiring model retraining. 
To structure the reasoning process, the CoT component guides the LLM through a sequential path via ``Analyze $\Rightarrow$ Retrieve $\Rightarrow$ Assign $\Rightarrow$ Sequence $\Rightarrow$ Generate Code''. 
We further reinforce this by adopting a ``Code-as-CoT'' paradigm, where a demonstration in the prompt conditions the model to express its final plan as executable code. 
This method enforces logical consistency and improves the reliability of the generated plans.

This entire inference process shapes from constructing the prompt to generating the final Solution Package is described in Algorithm $1$ (see Appendix). 
The Solution Package integrates a natural language summary, the structured thought process, and the auditable machine instructions, providing a transparent basis for subsequent human-in-the-loop verification.

\subsection{Closed-Loop Verification and Execution} 
This final module embeds a closed-loop, Human-in-the-Loop (HIL) verification process to ensure the LLM-generated mission is safe and reliable. 
This critical step serves a dual purpose, as it acts as a safeguard against logical or factual errors in the LLM's reasoning and then incorporates essential human judgment to align the static plan with the unpredictable, dynamic conditions of the real-world operational environment.
This module implements a ``propose-and-confirm'' interaction model, prioritizing radical transparency to facilitate informed human oversight. 
Upon receiving the Solution Package, the interface presents to the operator with a three-fold view, including $1$) a concise natural language summary for immediate comprehension underlying the on-site environment, $2$) the complete CoT rationale, available on-demand for traceability, and $3$) the auditable, low-level, executable code to be implemented. 
By providing these insights, the system redefines the operator's role, transforming them into a real-time monitoring and decision-making authority capable of validating the plan at both a strategic and an implementation level.

When the operator rejects a plan based on their domain expertise or real-time situational awareness unavailable to the model (e.g., a sudden gust of wind not present in the static knowledge base, or an inefficient scan pattern chosen by the model for the current terrain), their corrective feedback is treated as a high-priority constraint. 
This feedback triggers a re-planning cycle, prompting the Manager Agent to generate a new solution that adheres to the revised constraints.
The framework proceeds to the execution stage only upon explicit operator confirmation.
At this point, the executable machine instructions from the Solution Package are passed directly to a secure execution environment (e.g., the simulator or a real UAV's API endpoint). 
This ensures a deterministic and verifiable transition from the audited code to physical actuation, thereby completely mitigating the risk of LLM-introduced errors during execution.

\section{Experiments and Results}
We evaluate the proposed LLM-Cognitive Reasoning Framework through a series of rigorously designed experiments based on a complex disaster response mission. 
The evaluation quantitatively compares the framework against baseline methods across three perspectives: mission success rate and planning quality, operator cognitive workload, and robustness to dynamic uncertainties.

\subsection{Experimental Setup}

\noindent\textbf{Simulation Environment \& Scene Elements \& UAV Swarm Configuration.}
With the ethics approval obtained from our institution, this study recruited $10$ UAV operators with varying levels of experience, specifically with $3$ experts, $4$ intermediate, and $3$ novices, for task implementation and NASA-TLX self-reporting.
All experiments were conducted under the same computational environment, which is AirSim \cite{shah2017aerial}, a high-fidelity simulator built on Unreal Engine $4$. 
To ensure experimental diversity and reproducibility, we developed a unified, parameterized disaster scene generator that randomly creates 10 distinct site scenarios within a $2 km \times 2 km$ area. 
Each participant attempted all $4$ comparative methods across all $10$ scenarios, yielding a total of $400$ experimental trials.
The core elements and their parameter distributions for each scene are detailed in Table \ref{tab:scene_elements_final}.

% Table for Scene Elements
\begin{table}[hbt!]
\centering
\caption{Main Parameters for Randomized Disaster Scene Generation.}
\label{tab:scene_elements_final}
\resizebox{\textwidth}{!}{
\begin{tabular}{lcll} 
\toprule
\textbf{Element}                        & \textbf{Quantity}     & \textbf{Scale / Intensity Parameter}        & \textbf{Constraint / Note}                           \\ 
\midrule
\multirow{2}{*}{\textbf{Base Station}}  & \multirow{2}{*}{$1$}    & \multirow{2}{*}{Coordinate: $[0, 0, 0]$}    & The origin for the swarm and the com-                \\
                                        &                       &                                             & munication anchor for the Relay UAV.                 \\ 
\hline
\multirow{3}{*}{\textbf{Disaster Zone}} & \multirow{3}{*}{$1$}    & Radius: \SI{500}{\meter}                    & All other elements (Obstacles, Surviv-               \\
                                        &                       & Centered at a random location               & ors) are procedurally spawned within                 \\
                                        &                       & $>$\SI{600}{\meter} from Base Station       & ~this zone.                                          \\ 
\hline
\multirow{3}{*}{\textbf{Obstacles}}     & \multirow{3}{*}{$5$–$10$} & Type: Cube, Cylinder, Wall                  & Static obstacles. Any collision results~in           \\
                                        &                       & Height: \SIrange{10}{45}{\meter}            & immediate mission failure. The model                 \\
                                        &                       &                                             & must learn their positions via mapping.              \\ 
\hline
\multirow{2}{*}{\textbf{Survivors}}     & \multirow{2}{*}{$1$–$5$}  & Point source                                & Static heat sources. Primary targets for~            \\
                                        &                       & Thermal Signature:~$0.8$ - $1.0$                & the Searcher UAV.                                    \\ 
\hline
\multirow{3}{*}{\textbf{Wind Zones}}    & \multirow{3}{*}{$0$–$2$}  & Spherical volume                            & Dynamic Event: Appears after~$t > \SI{60}{\second}$  \\
                                        &                       & Radius: \SIrange{50}{100}{\meter}           & at a random location. Entry into this~               \\
                                        &                       & Vector: \SIrange{10}{15}{\meter\per\second} & zone results in mission failure.                     \\
\bottomrule
\end{tabular}
}
\end{table}

The UAV swarm setting comprises three heterogeneous AirSim quadrotors. 
The key static properties of each agent, including functional designation, sensor suite, and operational parameters, are provided to the LLM and detailed in Table \ref{tab:drone_config_final}.

% Table for UAV Configuration
\begin{table}[hbt!]
\centering
\caption{UAV Swarm Configuration Parameters.}
\label{tab:drone_config_final}

\begin{tabular}{lcll} 
\toprule
\textbf{ID}                     & \textbf{Max. Speed}                         & \textbf{Role /~\textbf{Sensor Payload}} & \textbf{Primary Duty}                      \\ 
\midrule
\multirow{3}{*}{\textbf{UAV-$1$}} & \multirow{3}{*}{\SI{10}{\meter\per\second}} & Inspector                               & Performs global mapping at high~           \\
                                &                                             & ($ImageType.Scene$)               & altitudes ($h \in [50, 100]~m$) to provide~  \\
                                &                                             &                                         & obstacle data for the swarm.               \\ 
\midrule
\multirow{3}{*}{\textbf{UAV-$2$}} & \multirow{3}{*}{\SI{10}{\meter\per\second}} & Searcher                                & Conducts low-altitude search~              \\
                                &                                             & ($ImageType.Infrared$)            & ($h \in [10, 30]~m$) to detect and~          \\
                                &                                             &                                         & localize survivor heat signatures.         \\ 
\midrule
\multirow{3}{*}{\textbf{UAV-$3$}} & \multirow{3}{*}{\SI{10}{\meter\per\second}} & Relay                                   & Maintains a continuous com-~               \\
                                &                                             & (Communications Package)                & munication link between the~               \\
                                &                                             &                                         & agents and the base station.               \\
\bottomrule
\end{tabular}
\end{table}

\vspace{2mm}
\noindent\textbf{Mission Setup \& Baseline Comparisons.}
We designed a unified, multi-objective disaster response mission, which requires all evaluated methods to generate a single, directly executable Python script. 
This script must autonomously coordinate three parallel objectives, including $1$) \textit{Mapping}: UAV-$1$ surveys the disaster zone to map obstacle locations, $2$) \textit{Searching}: UAV-$2$ performs low-altitude infrared scanning to locate survivors, $3$) \textit{Relaying}: UAV-$3$ maintains a continuous communication link.
The mission is governed by the following rigorous constraints, where any one of the following violation results in immediate failure.
Collision avoidance, that is, no UAV collides with obstacles, UAV-$2$ must maintain an altitude of $h > 50~m$ while in unmapped areas.
Communication Link requires the distance between UAV-$1$ -$2$ and UAV-$3$ must not exceed $400m$, and the distance between UAV-$3$ and the base station must be $\leq 1000~m$.
Dynamic Hazard Evasion suggests that no UAV allows to enter an active wind zone.
All methods receive identical inputs: the initial site state (disaster zone location and UAV initial states) and complete API documentation. 
Specifically, we developed four configurations for experimental evaluation:
\begin{itemize}[leftmargin=*, itemsep=0pt]
    \item \textbf{B$1$ (Manual)}: Human operators manually write, debug, and execute mission code using Python and AirSim API.
    \item \textbf{B$2$ (LLM-Direct)}: Base Qwen-$14$B-Chat model without the proposed curated Static Domain Knowledge Base or Chain-of-Thought examples.
    \item \textbf{B$3$ (Ours w/o Feedback)}: Our system with ICL yet without human-in-the-loop feedback and executes the first generated plan without operator confirmation.
    \item \textbf{Ours (Full, LLM-CRF)}: The proposed complete LLM-CRF with ICL and human feedback loop.
\end{itemize}

% \vspace{1mm}
\noindent\textbf{Evaluation Metrics.}
Since this work is the first of its kind to adopt an LLM in disaster SAR tasks, there is no established benchmarks for comparison.
With the objective to fairly evaluate the proposed LLM-CRF, this study establishes quantitative metrics and compares its performance against baseline approaches across three dimensions, including mission success rate, task quality, and operational efficiency (i.e., operator's cognitive workload).

\textit{Mission Success Rate (MSR)}: For each trial $j$, $MSR_j = 1$ if all objectives are met without constraint violations, else $MSR_j = 0$. Overall success rate is $MSR = \frac{1}{N}\sum_{j=1}^{N} MSR_j \times 100\%$.

\textit{Task Quality (TQ)}: For all trials (including partial completions in failures), we measure Search Coverage ($Cov_{search}$) as the percentage of disaster area scanned by UAV-$2$'s sensor footprint ($Cov_{search} = A_{covered}/A_{total} \times 100\%$), and Survivors Found ($Rate_{found}$) as the percentage of survivors correctly located ($Rate_{found} = N_{found}/N_{total} \times 100\%$).

\textit{Efficiency \& Operator Load}: Total Mission Time (TMT) is recorded in seconds from task start to completion (or failure). 
Cognitive workload is measured through the NASA Task Load Index (NASA-TLX) \cite{hart1988development}, where operators rate their task experience on six dimensions (i.e., Mental Demand, Physical Demand, Temporal Demand, Performance, Effort, Frustration) using a $21$-point scale (rated from $0$–$100$, the lower score indicates the less cognitive load the operator has), then the complete pairwise comparisons to weight these dimensions. 
The final weighted score is:
\begin{equation}
\label{eq:nasa-tlx}
    Load_{TLX} = \frac{1}{15} \sum_{i} R_i \cdot w_i, \quad i \in \{\text{MD, PD, TD, Perf, Eff, Frus}\}
\end{equation}

\subsection{Experimental Results Analysis}

We conducted experiments with $10$ participants, each of them attempted all $4$ baselines across $10$ randomly generated disaster scenarios, yielding a total of $400$ experimental trials. 
The main results are summarized in Table \ref{tab:main_results}.

\begin{table}[hbt!]
\centering
\caption{Averaged Experimental Results on the Unified Mission.}
\label{tab:main_results}
\resizebox{\textwidth}{!}{
\begin{tabular}{lcccc}
\toprule
\textbf{Metric}           & \textbf{B1 (Manual)} & \textbf{B2 (LLM-Direct)} & \textbf{B3 (Ours w/o Feedback)} & \textbf{Ours (Full)}     \\ 
\midrule
Mission Success Rate (\%) & 87.0                 & 11.0                     & 62.0                            & \textbf{94.0}            \\ 
\midrule
Search Coverage (\%)      & 94.8 $\pm$ 4.2       & 71.3 $\pm$ 19.8          & 92.3 $\pm$ 4.8                  & \textbf{96.2} $\pm$ 2.8  \\
Survivors Found (\%)      & 93.1 $\pm$ 3.9       & 68.5 $\pm$ 21.3          & 79.8 $\pm$ 14.6                 & \textbf{94.8} $\pm$ 3.1  \\
Total Mission Time (s)    & 1295 $\pm$ 418       & 393 $\pm$ 287            & \textbf{387} $\pm$ 42           & 463 $\pm$ 51             \\
NASA-TLX Score (\%)           & 71.2 $\pm$ 9.3       & 68.5 $\pm$ 13.7          & 42.8 $\pm$ 8.1                  & \textbf{28.3} $\pm$ 6.2  \\
\bottomrule
\end{tabular}
}
\end{table}

\vspace{1mm}
\noindent\textbf{Overall Performance Analysis.}
As shown in Table \ref{tab:main_results}, the proposed LLM-CRF system (Ours) demonstrates substantial superiority across all core metrics. 
It achieved a $94.0\%$ Mission Success Rate, highlighting its robustness in complex, long-horizon planning under multiple constraints. 
In contrast, B$2$ (LLM-Direct), which lacks domain knowledge and structured reasoning, failed in most cases, of only $11.0\%$ success rate obtained. 
This might be due to the fact that the generated code from this approach generally contains logical errors or fails to address implicit dependencies (e.g., mapping before low-altitude flight).

Regarding task quality among successful runs, the proposed LLM-CRF slightly outperformed the B$1$ (Manual), achieving a higher Survivors Found Rate ($94.8\%$ vs. $93.1\%$) with significantly lower variance ($3.1\%$ vs. $3.9\%$). 
More critically, the LLM-CRF demonstrated superior Search Coverage ($96.2\%$ vs. $94.8\%$), indicating more systematic and exhaustive search patterns. 
Analysis of failed cases revealed that human operators, though capable, were susceptible to planning fatigue, which led to suboptimal scan paths with coverage gaps.

In terms of efficiency and cognitive workload, the LLM-CRF system reduced the average mission time by $64.2\%$ compared to manual coding ($463s$ vs. $1295s$) and lowered the NASA-TLX score by $42.9\%$ ($28.3$ vs. $71.2$), effectively shifting operators from a high to a low cognitive load condition.
Although it significantly reduces mental workload, the proposed LLM-CRF design did not exclude the operator personnel from the operational loop or their decision-making role. 
Instead, it underscores the human's critical responsibilities in monitoring and inference, which significantly enhances its potential for future deployment.

\noindent\textbf{Task Complexity and the Critical Role of Human Feedback.}
A particularly revealing comparison exists between the full LLM-CRF system and the ablated version of B$3$ (Ours without Feedback). 
Specifically, B$3$ achieved the fastest execution time of $387$ seconds, and demonstrated strong performance on simple subtasks. 
For instance, its Search Coverage of $92.3\%$ was competitive with $96.2\%$ of the full version and the $94.8\%$ of B$1$. 
However, its performance deteriorated significantly on complex and safety critical tasks. 
The Survivors Found rate for B$3$ fell to $79.8\%$, with a high variance of $\pm14.6\%$, and its overall Mission Success Rate dropped sharply to $62.0\%$.
This result constitutes a $32\%$ absolute increase in the failure rate when compared to the full system, which achieved a $94.0\%$ success rate.
This performance gap highlights a critical insight, that is, \textit{the LLM-CRF system has already achieved human competitive competence on well-defined, static subtasks}, yet \textit{human oversight remains indispensable for handling dynamic uncertainties}. 
Analysis of B$3$ failures revealed that $38\%$ of missions failed due to:
\begin{itemize}[leftmargin=*, itemsep=0pt]
    \item Collision with dynamically emerging wind zones ($19\%$ of trials)
    \item Suboptimal relay station placement causing communication loss ($12\%$)
    \item Edge-case path-finding errors in complex obstacle fields ($7\%$)
\end{itemize}

\begin{figure}[t]
    \centering
    \includegraphics[width=\linewidth]{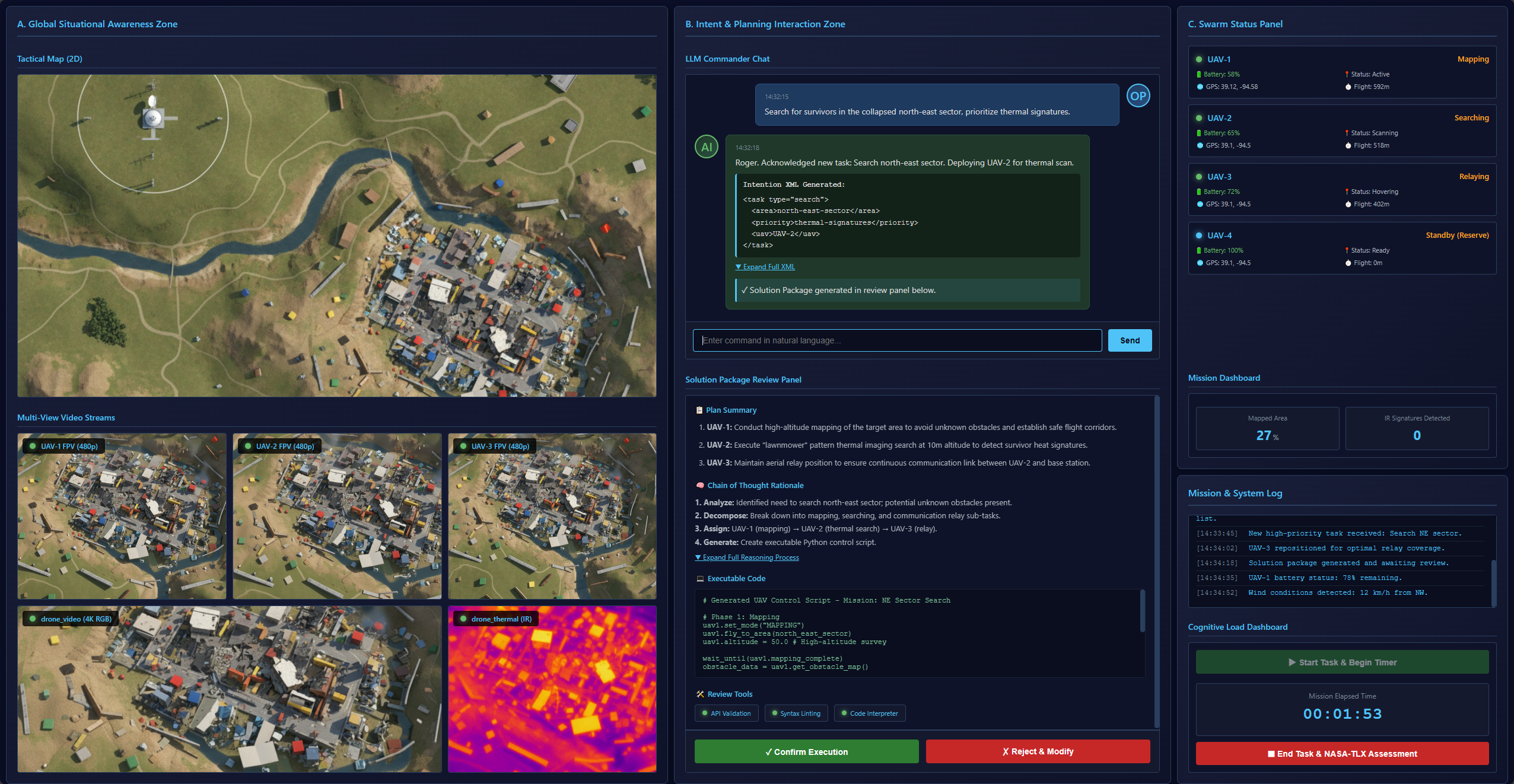}
    \caption{\textbf{The proposed LLM-CRF interface demonstration.}
    Left: Real-time UAV swarm site scenario (video stream and thermal feedback); Middle: LLM-CRF dialogue with generated plans; Right: Dashboard for UAV and task parameters.}
    \label{fig3}
\end{figure}

In contrast, when human operators reviewed the initial plans generated by the LLM-CRF and provided lightweight corrective feedback, the system successfully re-planned in $97\%$ of flagged cases. 
For example, an operator might following the instruction, ``Avoid the northwest quadrant'' upon receiving a weather alert. 
This result demonstrates that human-in-the-loop verification acts not merely as a safety net, but as a strategic necessity for bridging the gap between static reasoning and dynamic reality. 
This collaborative approach ensured a $51.6\%$ relative improvement in the success rate, raising it from $62.0\%$ to $94.0\%$, while introducing only a minimal cognitive burden, as reflected in the NASA-TLX score which increased from $28.3$ to $42.8$.
Figure \ref{fig3} demonstrates the interface for the proposed LLM-CRF as it generates task plans and operational commands.

\section{Conclusion}
This paper introduces the LLM-CRF system that enables an LLM to function as an autonomous mission commander for UAV swarms undet complex disaster response scenarios. 
The core of the proposed approach is a structured reasoning process that synergizes perceptual alignment, in-context learning, and closed-loop verification to transform a general-purpose LLM into a reliable planner, effectively bridging the gap between high-level reasoning and safe, embodied, real-case execution.
A central contribution of this work is its human-on-the-loop paradigm, which redefines the operator's role from manual coder to strategic supervisor. 
The framework grounds NLP commands in a real-time model, autonomously decomposes them into parallel sub-tasks (e.g., mapping, search, relay), and generates a transparent CoT rationale alongside the final executable code. 
This design ensures operational safety and logical soundness by mandating human validation before any action is taken.

Extensive experimental results provide robust quantitative validation of the proposed approach. 
Specifically, the LLM-CRF achieved a $94.0\%$ Mission Success Rate under hard safety constraints, with a Search Coverage of $96.2\%$ and a Survivors Found Rate of $94.8\%$, demonstrating its competence in generating high-quality and safe plans. 
Crucially, this performance was also maintained with the operator's cognitive load (NASA-TLX) of $28.3\%$, confirming the framework's success in alleviating the mental burden of complex swarm management.

While this work provides a promising foundation for human-machine teaming in critical missions, its performance is contingent on high-quality sensor data, which yields a limitation of the current evaluation. 
Future degradation from significant sensor noise or failures remains a key challenge. 
Addressing this by integrating predictive environmental models and robust dialogue-based re-planning constitutes a vital direction for future work, essential for transitioning from simulation to real-world deployment. 
Through fusing LLM-based strategic reasoning with human oversight, this framework provides a prior study toward deploying autonomous and reliable robotic systems.

% \section*{References}
% References follow the acknowledgments in the camera-ready paper. Use unnumbered first-level heading for
% the references. Any choice of citation style is acceptable as long as you are
% consistent. It is permissible to reduce the font size to \verb+small+ (9 point)
% when listing the references.
% Note that the Reference section does not count towards the page limit.
\newpage
\medskip
\bibliographystyle{IEEEtran}
{
\small

\bibliography{reference}
}

%%%%%%%%%%%%%%%%%%%%%%%%%%%%%%%%%%%%%%%%%%%%%%%%%%%%%%%%%%%%

\appendix

\section{Technical Appendices and Supplementary Material}
\begin{algorithm}[H]
\caption{Swarm Task Planning via In-Context Learning with Chain-of-Thought}
\label{alg:planning_icl_cot}
\begin{algorithmic}[1]
\Statex \textbf{Input:} Intent $I$, World State $S$
\Statex \textbf{Global Context:} Knowledge Base $KB$, API Docs $API$, Exemplars $E = \{(e_i^{task}, e_i^{cot}, e_i^{code})\}$

\Function{GenerateSolutionPackage}{$I, S$}
    \State $Context_{domain} \gets \text{RetrieveKnowledge}(I, KB)$; $e_{sim} \gets \text{FindMostSimilarExemplar}(I, E)$
    \State $Analysis \gets \text{LLM-CoT}(\text{"Analyze intent"}, I, S, Context_{domain}, e_{sim}^{task}, e_{sim}^{cot})$
    \State $Tactics \gets \text{LLM-CoT}(\text{"Identify tactics"}, Analysis, KB.tactics, e_{sim}^{cot})$
    \State $T_{tree} \gets \textbf{DecomposeAndAssign}(I, Analysis, Tactics, KB, E)$
    \State $Plan_{seq} \gets \text{LLM-CoT}(\text{"Sequence tasks"}, T_{tree}, KB.constraints, e_{sim}^{cot})$
    \State $\Psi_{final} \gets \textbf{GenerateExecutableCode}(T_{tree}, Plan_{seq}, API, E)$
    \State \textbf{return} \text{Package}($\Psi_{final}$, $T_{tree}$, $Analysis$, $Plan_{seq}$)
\EndFunction

\Function{DecomposeAndAssign}{$I, Analysis, Tactics, KB, E$}
    \State $T_{root} \gets \{task: I, role: null, subtasks: [\,]\}$
    \State \textbf{return} \textbf{RecursiveDecompose}($T_{root}, Analysis, Tactics, KB, E$)
\EndFunction

\Function{RecursiveDecompose}{$T_{current}, Analysis, Tactics, KB, E$}
    \State $e_{ref} \gets \text{FindMostSimilarExemplar}(T_{current}.task, E)$
    \State $IsAtomic \gets \text{LLM-CoT}(\text{"Is atomic?"}, T_{current}.task, Tactics, KB.capabilities, e_{ref}^{cot})$
    \If{$IsAtomic = \text{True}$}
        \State $T_{current}.role \gets \text{LLM-CoT}(\text{"Assign role"}, T_{current}.task, KB.uav\_roles, e_{ref}^{cot})$
        \State \textbf{return} $T_{current}$
    \Else
        \State $Subtasks \gets \text{LLM-CoT}(\text{"Decompose"}, T_{current}.task, Tactics, KB.constraints, e_{ref}^{cot})$
        \For{each $st \in Subtasks$}
            \State $T_{child} \gets \{task: st, role: null, subtasks: [\,]\}$
            \State $T_{child} \gets \textbf{RecursiveDecompose}(T_{child}, Analysis, Tactics, KB, E)$
            \State $T_{current}.subtasks.\text{append}(T_{child})$
        \EndFor
        \State \textbf{return} $T_{current}$
    \EndIf
\EndFunction

\Function{GenerateExecutableCode}{$T_{tree}, Plan_{seq}, API, E$}
    \State $CodeFragments \gets \{\}$
    \For{each leaf $T_{leaf} \in \text{GetLeaves}(T_{tree})$}
        \State $uav\_id \gets T_{leaf}.role$; \quad $e_{code} \gets \text{FindCodeExemplar}(T_{leaf}.task, uav\_id, E)$
        \State $code_{leaf} \gets \text{LLM}(\text{"Generate code"}, T_{leaf}.task, API[uav\_id], e_{code}^{code})$
        \State $CodeFragments[T_{leaf}] \gets code_{leaf}$
    \EndFor
    \State $e_{asm} \gets \text{FindAssemblyExemplar}(T_{tree}, E)$
    \State $\Psi_{final} \gets \text{LLM}(\text{"Assemble script"}, T_{tree}, Plan_{seq}, CodeFragments, e_{asm}^{code})$
    \State \textbf{return} $\Psi_{final}$
\EndFunction

\Function{RetrieveKnowledge}{$I, KB$}
    \State $keywords \gets \text{ExtractKeywords}(I)$
    \State \textbf{return} $\{KB.tactics[k], KB.constraints[k] \mid k \in keywords\}$
\EndFunction

\Function{FindMostSimilarExemplar}{$task, E$}
    \State $scores \gets [\text{Similarity}(task, e_i^{task}) \mid e_i \in E]$
    \State \textbf{return} $E[\text{argmax}(scores)]$
\EndFunction
\end{algorithmic}
\end{algorithm}

%%%%%%%%%%%%%%%%%%%%%%%%%%%%%%%%%%%%%%%%%%%%%%%%%%%%%%%%%%%%

\end{document}